\documentclass[10pt,a4paper,conference]{IEEEtran}
\usepackage{amsmath,graphicx}
\usepackage{tabularx,ragged2e}
\usepackage{subfig}
\usepackage{csquotes}
\usepackage{arydshln}
\usepackage{caption}
\usepackage{booktabs}
\usepackage{cite}
\UseRawInputEncoding
\usepackage{color}

\newcommand{\etal}{\textit{et al}.}
\newcommand{\ie}{\textit{i}.\textit{e}.}
\newcommand{\eg}{\textit{e}.\textit{g}.}
\newcommand{\etc}{\textit{etc}}

\ifCLASSINFOpdf
\else
\fi
\begin{document}
%

\title{Feature-Supervised Action Modality Transfer}

\author{
\IEEEauthorblockN{Fida Mohammad Thoker }
\IEEEauthorblockA{University of Amsterdam\\
f.m.thoker@uva.nl}
\and
\IEEEauthorblockN{Cees G. M. Snoek}
\IEEEauthorblockA{University of Amsterdam\\
cgmsnoek@uva.nl}
 
}


%


\maketitle

\begin{abstract}
%
This paper strives for action recognition and detection in video modalities like RGB, depth maps or 3D-skeleton sequences when only limited modality-specific labeled examples are available. For the RGB, and derived optical-flow, modality many large-scale labeled datasets have been made available. They have become the de facto pre-training choice when recognizing or detecting new actions from RGB datasets that have limited amounts of labeled examples available. Unfortunately, large-scale labeled action datasets for other modalities are unavailable for pre-training. In this paper, our goal is to recognize actions from limited examples in non-RGB video modalities, by learning from large-scale labeled RGB data. 
To this end, we  propose a two-step training process: (\textit{i}) we extract action representation knowledge from an RGB-trained teacher network and adapt it to a non-RGB student network. (\textit{ii}) we then fine-tune the transfer model with available labeled examples of the target modality. For the knowledge transfer we introduce feature-supervision strategies, which rely on unlabeled pairs of two modalities (the RGB and the target modality) to transfer feature level representations from the teacher to the student network. 
Ablations and generalizations with two RGB source datasets and two non-RGB target datasets demonstrate that an optical-flow teacher provides better action transfer features than RGB for both depth maps and 3D-skeletons, even when evaluated on a different target domain, or for a different task. Compared to alternative cross-modal action transfer methods we show a good improvement in  performance especially when labeled non-RGB examples to learn from are scarce.
\end{abstract}


%
\IEEEpeerreviewmaketitle

\section{Introduction}
The goal of this paper is to recognize an action like \textit{drinking water}, \textit{hugging} or \textit{falling down} in multimodal video content, be it a stream of RGB pixels~\cite{slow_fast,kinetics,hara3dcnns}, depth maps \cite{Depthmotionmap,3ddepth,DBLP:deep_depth} or 3D-skeletons \cite{2sagcn2019cvpr,Shi_2019_CVPR,Wu_2019_ICCV}. The common approach to action recognition in video is to train a deep convolutional neural network on massively labeled RGB, or derived optical-flow, video datasets like Kinetics~\cite{kinetics}, 
Sports-1M~\cite{Sports1M} or ActivityNet~\cite{caba2015activitynet}. These pre-trained RGB models are also valuable to recognize or detect new actions from alternative RGB videos, with only limited amounts of labeled action examples available for fine-tuning \cite{ucf-101,hmdb,THUMOS14}, thereby saving a lot of annotation cost. Unfortunately, for non-RGB video modalities massively labeled action datasets, and the corresponding pre-trained models, are scarce. In this paper, we strive for limited-example action recognition in non-RGB video modalities, like depth maps and 3D-skeletons, by learning from large-scale RGB video data labeled with other actions.

We take inspiration from the ideas of general knowledge distillation by Hinton \etal~\cite{kd}  and cross-modal distillation for action recognition. \eg~\cite{garci_md,garcia2018admd,crasto2019mars,fmthoker_icip}. The goal of knowledge distillation is to compress a large complex teacher network into a small and simple student network. In cross-modal distillation a teacher network is first trained to recognize a set of actions from the source action modality using many labeled examples. Then the teacher network  distills knowledge to the student network to recognize the same set of actions from a different target action modality. We adapt these  ideas for a different setting. That is, to train a student network to recognize a set of actions from a target modality while distilling knowledge from a teacher network that has been trained to recognize a different set of actions from a different source action modality. This scenario has a practical application for recognizing or detecting new actions from non-RGB action modalities with limited labeled examples. Instead of relying on labeled large scale datasets of these non-RGB modalities for pre-training (which are scarce), we can  distill knowledge from existing RGB trained action models. In summary, we aim to transfer information about recognizing actions across modalities and across classes.

In this paper, we propose to recognize and detect actions from  non-RGB modalities like depth maps and 3D-skeleton sequences,  when only limited labeled examples for these modalities are available. 
To achieve this, we assume an RGB trained action model is given and we also have access to many unlabeled pairs of two modalities (paired RGB and non-RGB actions), along with some labeled examples for the non-RGB action modality. Then, the trained RGB model acts as the teacher network to supervise the learning of the non-RGB student model using unlabeled modality pairs. In contrast to the general knowledge transfer, which distills class probabilities from the teacher to the student network, we distill action representations from the teacher to the student network via \textit{feature-level supervision}.  More precisely, for a given unlabeled modality pair, the non-RGB student network is optimized  to match its output features with that of the trained RGB teacher. After this cross-modal distillation step, the non-RGB student network is fine-tuned with the available labeled examples of the non-RGB modality for a downstream task. Before presenting our method, we will first discuss in more detail related work.

\section{Related Works}

\subsection{Modalities for Action Recognition }
Modern action recognition, 
\eg,~\cite{kinetics,hara3dcnns,c3d,r2d+1,TSN2016ECCV}  relies on deep (2D or 3D) CNN architectures that learn to classify human actions from video data. These methods usually require a common video modality such as RGB, the RGB-derived optical-flow or both~\cite{SimonyanZ14,feichtenhofer2016convolutional,zhao2019twoinone} to achieve best performance for this task. For these video modalities many large-scale and publicly available annotated action datasets exists, such as Kinetics-(400, 600 and 700) \cite{kinetics}, Sports-1M \cite{Sports1M}, and ActivityNet \cite{caba2015activitynet}. These sets also act as valuable pre-training resource for classifying and detecting new actions from other RGB action datasets, which have smaller amounts of labeled examples.

There is also a large body of works that learn to classify human actions from  other video modalities such as depth maps~\cite{Depthmotionmap,stop,3ddepth,DBLP:deep_depth}, sequences of 3D-skeletons~\cite{2sagcn2019cvpr,Shi_2019_CVPR,Wu_2019_ICCV,stgcn2018aaai}, and even radio frequencies~\cite{li2019rf}. Although action recognition networks for these modalities may perform well, they require a large number of labeled action examples from the target modality for training. In contrast, our method utilizes large-scale labeled datasets of the commonly available RGB modality to boost the performance on non-RGB modalities, especially when only limited amounts of non-RGB labeled examples are available.

\begin{figure*}[t]
\centering
\begin{minipage}[b]{0.9\linewidth}
  \centerline{\includegraphics[width=17cm,height=6cm]{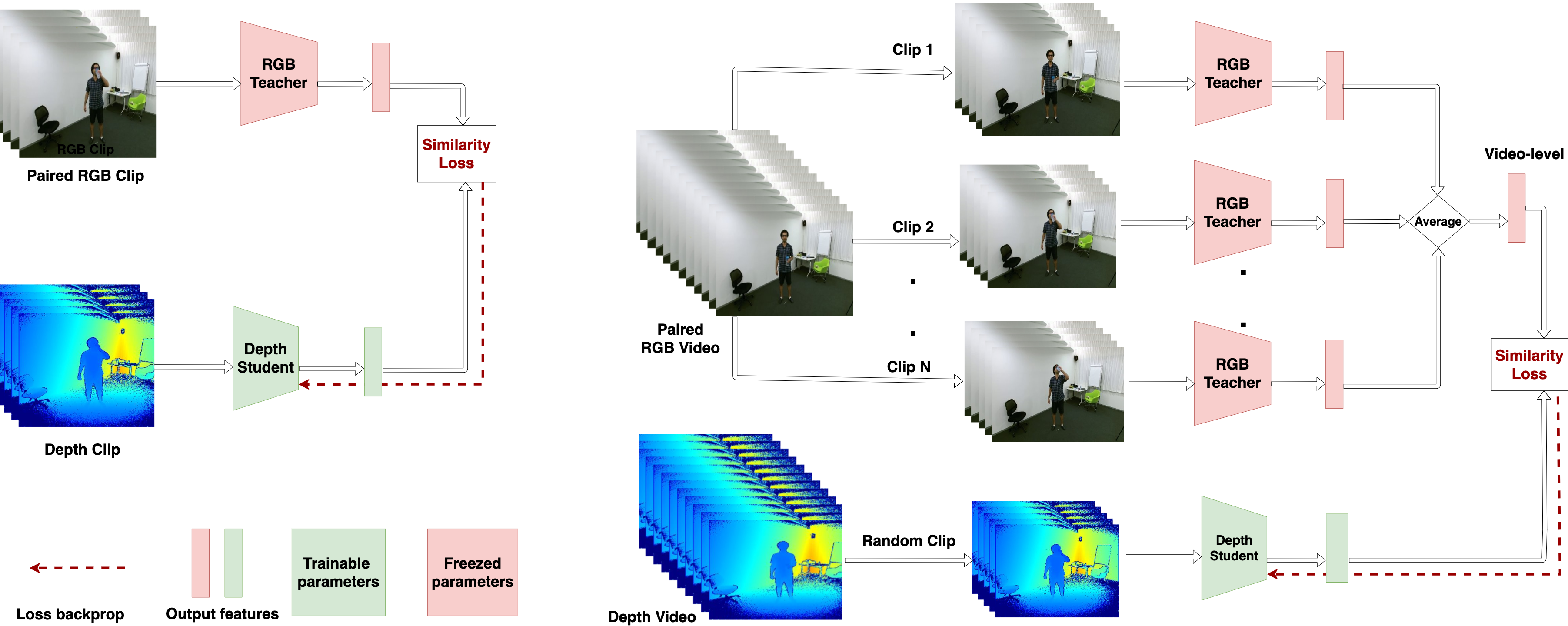}}
\end{minipage}
\caption{\textbf{\textit{Feature-supervised action modality transfer.}} A teacher network is trained on a large and labeled RGB, or derived optical-flow, action dataset. \emph{(left)} \textit{Clip-to-Clip transfer from the RGB teacher to the depth student.} A paired RGB-depth clip is sampled from the whole RGB-depth video pair and the student network is optimized to match its features from the depth clip with that of the teacher features obtained from the corresponding paired RGB clip. \emph({right)} \textit{ Video-to-Clip transfer from RGB to depth maps.} The whole RGB video is divided into $N$ clips to aggregate  clip level features from teacher network into a video-level feature. A clip is then randomly sampled from the paired whole depth video and the student network matches the clip features with the corresponding video level RGB features from the teacher. 
}
\label{fig:overview}
\end{figure*}

\subsection{Knowledge Transfer}
%
Recently, knowledge distillation has been explored to transfer knowledge across modalities for tasks like emotion recognition~\cite{emotion}, pose estimation~\cite{wall_pose}, object detection~\cite{hoffman2016learning,gupta2015cross,object_detect}, video captioning \cite{video_caption_1,video_caption_2} and action recognition~\cite{garci_md,garcia2018admd,crasto2019mars,d3d,garcia2019dmcl,fmthoker_icip,sayed:GCPR:2018}.

Gupta \etal \cite{gupta2015cross} transfer knowledge from the RGB to the depth modality for the task of object detection in images using a cross-modal teacher-student network that matches features between the two modalities. Similarly, Sayed \etal \cite{sayed:GCPR:2018} proposed a self-supervised method to learn feature similarity between RGB and optical flow modalities by maximizing similarity between clip features from paired RGB-flow videos and minimizing similarity across unpaired video clips. We rely on a similar principle, but different from both \cite{gupta2015cross} and \cite{sayed:GCPR:2018} we propose to exploit the temporal structure of video data for better information exchange via new feature-supervised granularities, like clip-to-clip, video-to-clip and video-to-video. These granularities not only improve transfer performance for action classification and detection, but are also necessary for challenging modality pairs like RGB and 3D-skeletons.

Thoker and Gall \cite{fmthoker_icip} proposed cross-modal transfer where a source (teacher) network is already trained to recognize a set of actions from the RGB modality. Their goal is to train a new (student) network to recognize the same set of actions, but from the skeleton modality.  Unlabeled  RGB-Skeleton action pairs are used such that the output action class predictions from the RGB teacher are matched by corresponding 3D-skeleton based student via common distillation losses like cross entropy (CE) or Kullback-Leibler divergence (KL).  We  deal with a more difficult variant of this problem, where the goal is to train a non-RGB student network to recognize a \textit{different} set of actions than those of the RGB teacher. Thus, the action classes seen by the teacher and the student network are disjunct for our case and the CE and KL losses used by Thoker and Gall are no longer applicable. Instead, we propose to rely only on feature-level supervision between teacher and student by minimizing the cosine distance. Instead of transferring class labels, we transfer action-specific feature-representations from the RGB modality to a non-RGB target modality. The features learned in this manner can be then fine-tuned to different downstream tasks.


Closest to our work are ~\cite{garci_md,garcia2018admd,crasto2019mars,d3d,garcia2019dmcl}, who all propose to extract knowledge from one or more source modalities to enhance action classification in a different target  modality. Particularly, both Crasto \etal~\cite{crasto2019mars} and Stroud \etal~\cite{d3d} boost the performance of RGB-only action recognition by distilling knowledge from a trained optical-flow teacher. Similarly, Garcia \etal~in \cite{garci_md} and \cite{garcia2019dmcl} rely on depth maps with or without optical-flow, as labeled paired source modalities to boost the performance of RGB-only action recognition. These methods also assume the network for the source modality is trained to recognize the same set of action classes as the target modality and they rely on labeled pairs (or triplets) of the respective modalities to transfer class-specific information from the source to the target modality. 
%
As mentioned previously, we assume the action classes of the source-based network to be different from  the target-based network and we rely on unlabeled pairs to transfer feature-level information from the source to the target modality.
As a result, our method can use large-scale labeled RGB action datasets as the pre-training data to recognize or detect  new  actions from non-RGB datasets with only a limited amount of labeled examples. 
We further differ from these methods by transferring knowledge to a more difficult target modality like 3D-skeleton data and extending cross-modal transfer to the task of temporal action detection.

\section{Proposed Method}
We consider the tasks of action classification and detection for a non-RGB modality, \eg~depth maps or sequences of 3D-skeletons, while requiring a reduced amount of labeled examples. To achieve this, a teacher-student network extracts knowledge from a pre-trained off-the-shelf RGB, or optical-flow, action model using unlabeled modality pairs. We first discuss our general approach for this feature-supervised knowledge transfer and then detail transfer granularities for different modality pairs.   

\subsection{Cross-Modal Teacher-Student}
Lets assume that we are given a network that has been trained on a large action-class labeled dataset of trimmed RGB videos. We call this dataset the source dataset and the corresponding network acts as the teacher network. We also assume to have access to another dataset called the target dataset that contains many unlabeled action pairs from two paired modalities --\ie, the RGB and the non-RGB target modality. The target dataset also contains some labeled action examples for the target modality, along with the unlabeled pairs. Note the action classes of the source and target dataset do not overlap. We train the student network to extract knowledge from the teacher network, which has learned from the labeled RGB modality of the source dataset, with the goal to adapt it to the non-RGB target modality. 

Formally, given a training pair ($V_{\textit{RGB}},V_\textit{{Target}}$) from unlabeled target data, the trained teacher network outputs a feature vector $F_\textit{{RGB}}$  for the RGB video $V_{\textit{RGB}}$. The student network uses the target modality $V_\textit{{Target}}$  as its input and is supervised by the corresponding RGB feature $F_\textit{{RGB}}$ from the teacher network. The student network is optimized to match its features $F_\textit{{Target}}$ with that of the teacher network using an appropriate similarity loss. We select for $F_{\textit{RGB}}$ and $F_{\textit{Target}}$ the outputs of the layer just before the fully connected layers of the teacher and the student network.
By doing so, we teach the student network to learn high-level semantics of the actions learned by the teacher, rather than learning the class-specific information present in the fully connected layers. Note that the student network can have same or different architecture as that of the teacher network, but the dimension of its output features   $F_{\textit{Target}}$  should be matched with $F_{\textit{RGB}}$, if different.
After the knowledge transfer step, the student network can be fine-tuned for a downstream task using the labeled data from the target dataset. During fine-tuning, a standard cross-entropy loss is used for the action classification and a regression loss for the temporal action detection. 

\subsection{Feature-Supervised Granularity}
The input granularity to the action recognition models depends on the  architecture of the network and the nature of the modality involved. For example, most state of the art works for image based modalities (RGB, optical-flow, depth, \etc.) use a small video clip as their input for predicting the action classes, while skeleton based models rely on a whole video sequence as the input. Similarly, during inference clip-level models combine prediction of different clips from the whole video to produce more accurate results. Thus, in order to explore the architecture of our cross-modal framework for different pairs of modalities and how to effectively extract knowledge from the teacher network, we consider three different transfer granularities, \ie, clip-to-clip, video-to-clip and video-to-video.

\subsubsection{Clip-to-Clip} We use this granularity for transfer between RGB and depth pairs, since both modalities require a small video clip as input. The transfer strategy for this granularity is shown in Fig.~\ref{fig:overview} (left). In particular, a paired small clip is sampled from the whole video of two modalities and a depth-based student network is optimized to match the features of the corresponding RGB clip. Thus, the student network learns to mimic the clip-level features of the RGB teacher by minimizing the following loss function:
 \begin{equation} \label{clip_loss}
\mathcal{L}\left(F_{\textit{Target}}^{\textit{clip}},F_\textit{{RGB}}^\textit{{clip}}\right) = \textit{Cosine\_Distance}\left(F_{\textit{Target}}^{\textit{clip}},F_\textit{{RGB}}^\textit{{clip}}   \right)
\end{equation}
where $F_{\textit{RGB}}^{\textit{clip}}$ and $F_{\textit{Target}}^{\textit{clip}}$ are the clip level features predicted by the RGB and depth networks respectively.

\subsubsection{Video-to-Clip} Different from the clip-to-clip transfer, the teacher network  extracts for this granularity information from the whole RGB video and then transfers it to the student network. This transfer strategy is shown in Fig.~\ref{fig:overview} (right). This strategy again fits for knowledge transfer between RGB and depth pairs. More  formally, the RGB  video is divided into $N$ clips ${C_1,C_2,...,C_N}$ of equal duration. For each clip $C_i$ the teacher network outputs a feature vector 
$\mathcal{F}_i$ and a global video level feature is obtained by combining these clip level features. Then, the student network randomly samples one of the clips from the corresponding paired depth video and uses the video level RGB feature from the teacher for supervision. Thus, the student network is  optimized to match each target clip with the corresponding video-level RGB feature from the teacher network by minimizing the following loss:
\begin{equation}
\label{video_to_clip_loss}
\mathcal{L}\left(F_{\textit{Target}}^{\textit{clip}},F_\textit{{RGB}}^\textit{{video}}\right) = \textit{Cosine\_Distance}\left(F_{\textit{Target}}^{\textit{clip}},F_\textit{{RGB}}^\textit{{video}}   \right)
\end{equation}
\begin{equation}
\label{}
F_{\textit{RGB}}^{\textit{video}} =  \frac{1}{N}\sum_{1}^{N}  \mathcal{F}_{i}^{\textit{clip}}
\end{equation} 
where $F_{\textit{RGB}}^{\textit{video}}$ is computed by taking an average over $N$ RGB clips. The number of clips $N$ for each video is calculated as $\frac{\textit{Total Number of frames }}{ \textit{Input size of the network}}$. 

Naturally, we can also combine the clip-to-clip and video-to-clip granularity, such that both clip-level and video-level information is transferred to the student network. In this scenario, the student network will be optimized to match features of each target clip with that of the corresponding paired RGB clip and 
with the global video level feature from the whole paired RGB video. The following loss function is minimized for this optimization: 
\begin{equation} \label{combined}
\mathcal{L}\left(F_{\textit{Target}}^{\textit{clip}},F_\textit{{RGB}}^\textit{{clip}}, F_\textit{{RGB}}^\textit{{video}}\right) =  \mathcal{L}\left(F_{\textit{Target}}^{\textit{clip}},F_\textit{{RGB}}^\textit{{clip}}\right) + \mathcal{L}\left(F_{\textit{Target}}^{\textit{clip}},F_\textit{{RGB}}^\textit{{video}}\right)  
\end{equation} 

We will assess the knowledge transfer abilities of these strategies in the ablation experiments.

\subsubsection{Video-to-Video} \label{subsec:vtov}
This granularity is required to transfer knowledge from the RGB to the 3D-skeleton modality, since the input to a skeleton-based network is the whole skeleton sequence. The  transfer strategy is the same as for the video-to-clip; the only change being the student network is now replaced by a skeleton based architecture whose input is the whole skeleton video, instead of a small clip. Now, the student network is  optimized to match the whole skeleton sequence with the global video-level feature of the paired RGB teacher by minimizing the following loss:
\begin{equation}
\label{video_to_video_loss}
\mathcal{L}\left(F_{\textit{Target}}^{\textit{video}},F_\textit{{RGB}}^\textit{{video}}\right) = \textit{Cosine\_Distance}\left(F_{\textit{Target}}^{\textit{video}},F_\textit{{RGB}}^\textit{{video}}   \right)
\end{equation}

where $F_{\textit{Target}}^{\textit{video}}$ is the output feature of the student network which operates over a whole sequence.

\section{Experimental Setup}

 \begin{figure}[]
  \centering
  \begin{tabular}{c }
\includegraphics[width=6.6cm,height=1.85cm]{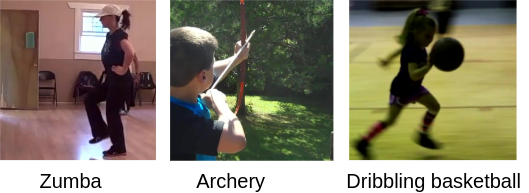}
\\
    (a) 
  \end{tabular}
  \begin{tabular}{c c}
\includegraphics[width=6.5cm,height=1.85cm]{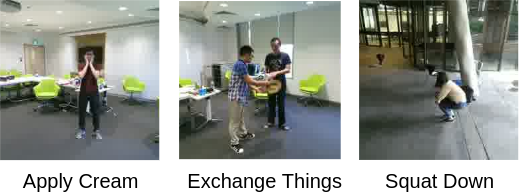} \\
    (b)
  \end{tabular}
    \begin{tabular}{c c c}
\includegraphics[width=6.5cm,height=4.8cm]{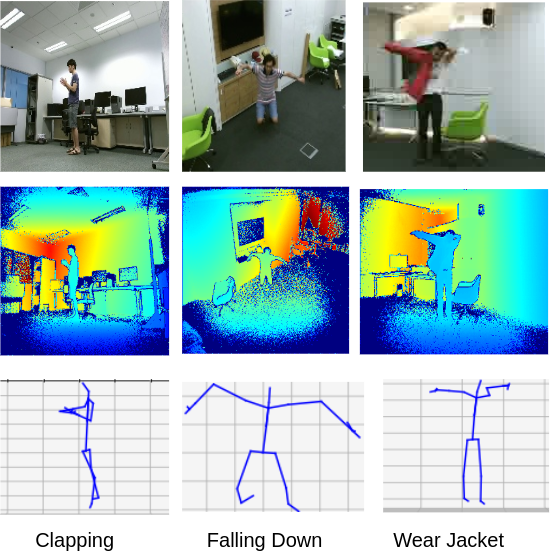} \\
    (c)
  \end{tabular}
  \caption{\textbf{\textit{Action modality samples from different source and target datasets}}. (a) Sample frames from the Kinetics-400 dataset. (b) Sample frames from the NTU-120\_minus\_60 dataset. (c) Paired multimodal samples from the NTU RGB+D  dataset. Note the domain change between the dataset in (a) and the other two datasets.} 
\label{fig:datasets}
\end{figure}
 
\subsection{Target Datasets}  
\textbf{NTU RGB+D}~\cite{Shahroudy_2016_NTURGBD} contains 56,880 trimmed videos with paired RGB, depth and 3D-skeleton modalities for 60 action classes. All actions are captured in indoor scenes with multiple cameras, different backgrounds, multiple subjects and camera setups. The dataset is split by a cross-view setup 
into 37,920 training and 18,960 validation videos \cite{Shahroudy_2016_NTURGBD}. For the knowledge transfer, we use the video modality pairs from the training set without any action class labels. For fine-tuning, we sample the action class examples of the target modality from the training set. The validation set is only used for evaluation purposes and is not seen during any training.

\textbf{PKU-MMD}~\cite{liu2017pku} contains 1,074 long untrimmed videos with paired RGB, depth and 3D-skeleton modalities for 51 action classes and about 20K action instances. All actions are captured in indoor scenes as well, with multiple cameras, different backgrounds, different views and 66 subjects. The annotations for each video contain the start and end locations of multiple activities, along with the action class. The dataset is split by a cross-subject setup 
into 942 training and 132 validation videos. We sample the labeled videos from the training set for the task of temporal action detection, for fine-tuning only. The validation set is used only for evaluation.

\subsection{Source Datasets}  
\textbf{Kinetics-400}~\cite{kinetics} is a large RGB-only dataset containing 400K labeled examples for 400 different action classes. The dataset is collected from YouTube with videos coming from a variety of sources, setups, \etc. Note the domain difference between this dataset and our target datasets is considerable.

\textbf{NTU-120\_minus\_60} Starting from the 
NTU RGB+D 120 \cite{Liu_2019_NTURGBD120} dataset, we remove all the videos that contain action classes overlapping with NTU RGB+D. Thus, this dataset shares the same domain as our target dataset, but contains different action classes. We only use the RGB modality of this dataset to train the teacher network.

We provide some target and source dataset samples in Fig.~\ref{fig:datasets}.

\subsection{Implementation Details} \label{subsec:implementation}
\textit{Teacher Network:} We use  3D-ResNets~\cite{hara3dcnns} with RGB or optical-flow as input for our teacher networks. For Kinetics-400, we rely on the pre-trained 3D-ResNext101 models available from \cite{crasto2019mars} and for NTU-120\_minus\_60 we train a 3D-ResNet18 from scratch. We use 16-frame clips as the input and all other hyperparameters from \cite{hara3dcnns}.

\textit{Depth Student Network:} For the \emph{depth maps}, we use a 3D-ResNet18 architecture as the student network with 16-frame clips as the input. We first apply a multi-scale corner cropping to the 16-frame clip as mentioned in \cite{hara3dcnns}, followed by  a resizing operation into a 3 x 16 x 112 x 112 sample and a horizontal flipping with 50\% probability. The network is  trained with an SGD optimizer using a weight decay of 0.001, 0.9 momentum, an initial learning rate of 0.1 and a batch size of 128. During cross-modal transfer we train the whole network for 400 epochs. For fine-tuning, we only train the fully connected layer and the last ResNet block for a total of 100 epochs. For evaluation, the predictions of all clips from the whole video are averaged for final classification.

\textit{3D-skeleton Student Network:} For the \emph{3D-skeleton data}, we use a Spatio Temporal Graph Convolution Network~\cite{stgcn2018aaai} as the student network with whole 3D-skeleton sequences as the input. Following the setting of \cite{stgcn2018aaai} no data augmentation is applied. The network is trained with an SGD optimizer using a weight decay of 0.00001, momentum 0.9, an initial  learning  rate  of 0.1 and a batch size of 100.  We train the network for 120 epochs during cross-modal transfer and 70 epochs during fine-tuning. During both steps, all layers of the ST-GCN network are trained. For evaluation, a class is predicted for a given 3D-skeleton sequence.

\section{Results} 

\subsection{Ablation}  
For all ablation experiments we consider the task of classifying actions of NTU RGB+D  from depth maps using only a limited amount of labeled depth examples. An action classification network trained with RGB or optical-flow modality from  the NTU\_120\_minus\_60 source dataset acts as the teacher network. For the cross-modal transfer step, all RGB-depth or optical-flow-depth pairs from the training set of NTU RGB+D are used without action labels as unlabeled modality pairs for knowledge transfer. For fine-tuning we sample some depth examples with action labels from the training set of NTU RGB+D (20, 50 or 100 examples per action class). For evaluation, we report the video-level accuracy on the  validation set from the cross-view split of NTU RGB+D. Each experiment is repeated five times with different random seeds and mean accuracy with variance is reported.

\begin{table}[]
 \centering
 \begin{tabular}{lccc}
 \toprule
  & \multicolumn{3}{c}{\textbf{Target-Modality: Depth}}\\ \cmidrule(lr){2-4}
  \textbf{Source-Modality} & 20 per-class &50 per-class & 100 per-class \\
  \midrule
RGB&  	{62.85}\scriptsize{$\pm$0.5} & {66.01}\scriptsize{$\pm$0.6} &{68.64}\scriptsize{$\pm$0.3} \\ 
Flow&   {68.43}\scriptsize{$\pm$0.2} & {71.53}\scriptsize{$\pm$0.1} &{73.43}\scriptsize{$\pm$0.3} \\ 
Two-stream & {69.16}\scriptsize{$\pm$0.5} & {72.10}\scriptsize{$\pm$0.5} & {74.41}\scriptsize{$\pm$0.5} \\ 
 %
 \bottomrule
 \end{tabular}
 \caption{\textbf{\textit{Which source modality.}} RGB and optical-flow teacher networks trained on the NTU\_120\_minus\_60 dataset. The two-stream fuses the individual predictions. 
 The video-to-clip strategy is used for knowledge transfer.
 The optical-flow teacher provides the best individual features for knowledge transfer to depth maps, fusion improves results further.
} \label{table:rgb_vs_flow}
\end{table}

\subsubsection{Which source modality} We first consider which source modality to use in the teacher network for knowledge transfer. Table ~\ref{table:rgb_vs_flow} shows the results for RGB and optical-flow as teacher modalities. While both modalities provide good features for knowledge transfer, the optical-flow teacher performs better than the RGB teacher. We attribute this to the motion features that contain better cues about the action representations for transfer, as compared to the RGB teacher, which provides features that mainly model spatial action information. Naturally, we can also combine the two source modalities into a two-stream network by fusing the results of the RGB and optical-flow student streams during inference.  This further improves the action classification performance, at the expense of double compute and parameters. In summary, the optical-flow teacher provides the best individual features for knowledge transfer and we focus on the optical-flow as the main source modality in the remaining experiments, unless indicated otherwise. 
 
\begin{table}[]
 \centering
 \begin{tabular}{lccc}
 \toprule
   & \multicolumn{3}{c}{\textbf{Target-Modality: Depth}}\\ \cmidrule(lr){2-4}
\textbf{Loss-Function}   & 20 per-class &50 per-class & 100 per-class \\
 \midrule
MSE&   {65.69}\scriptsize{$\pm$0.3} & {69.27}\scriptsize{$\pm$0.5} &{71.65}\scriptsize{$\pm$0.4} \\ 
Cosine&   {68.43}\scriptsize{$\pm$0.2} & {71.53}\scriptsize{$\pm$0.1} &{73.43}\scriptsize{$\pm$0.3} \\ 
 \bottomrule
 \end{tabular}
 \caption{\textbf{\textit{Which loss function.}}
 MSE vs. cosine loss for feature-supervised knowledge transfer. For both, the teacher network is trained on the optical-flow modality from NTU\_120\_minus\_60. 
 The video-to-clip strategy is used for knowledge transfer.
 We obtain better results with a cosine loss. } 
 \label{table:loss_function}
 \end{table}
 
\subsubsection{Which loss function} Since the goal of the student network is to match the features of the teacher network, any similarity (or distance) function will work for this optimization. We just consider two common loss functions: the mean square loss and the cosine distance loss in Table~\ref{table:loss_function}. We observe the cosine loss to work better for our task. Thus, for all other experiments we choose the cosine loss to optimize the student networks.

\subsubsection{Which granularity} Next we evaluate the granularity of the feature-supervised transfer. From the results in Table~\ref{table:which_supervision} we observe the video-to-clip transfer works better than the clip-to-clip transfer, especially when only few labeled examples are available for fine-tuning.  This is expected, as the student network in the video-to-clip strategy is optimized to match the video-level features that aggregate global information about the action representations from the whole video. Hence, this strategy provides better supervision, as compared to clip-to-clip features that transfer only local information about the action from a small clip. We also observe that combining the two strategies achieves the best knowledge transfer. 
%
%
To summarize, the video-to-clip transfer works better than the clip-to-clip transfer and combining the two gives an additional improvement. For the rest of the experiments we use this granularity for the feature-supervised transfer, unless indicated otherwise.
  
\begin{table}[]
 \centering
 \begin{tabular}{lccc}
 \toprule
  & \multicolumn{3}{c}{\textbf{Target-Modality: Depth}}\\ \cmidrule(lr){2-4}
  \textbf{Granularity} & 20 per-class &50 per-class & 100 per-class \\
  \midrule
Clip-to-Clip&  	{64.80}\scriptsize{$\pm$1.0} & {70.30}\scriptsize{$\pm$0.4} &{72.92}\scriptsize{$\pm$0.5} \\ 
Video-to-Clip&   {68.43}\scriptsize{$\pm$0.2} & {71.53}\scriptsize{$\pm$0.1} &{73.43}\scriptsize{$\pm$0.3} \\ 
Combined & {69.16}\scriptsize{$\pm$0.2} & {73.60}\scriptsize{$\pm$0.1} & {76.24}\scriptsize{$\pm$0.3} \\ 
 %
 \bottomrule
 \end{tabular}
 \caption{\textbf{\textit{Which granularity.}} Combining clip and video level features from the teacher network, as detailed in Eq.\eqref{combined}),  acts as the best feature-supervision strategy. 
} \label{table:which_supervision}
\end{table}

\subsubsection{Which layers to transfer}  We also explore which layers from the teacher network are best suited for the transfer. In particular we tried to match the output features of each ResNet block of the teacher network with the corresponding ResNet block of the student network, along with the last layer as before. We found that matching these additional layers does not add anything significant to the transfer and relying on the last-layer only still provides the best results.

\subsection{Domain Generalization} 

Next, we evaluate the effect of training with different source domains and compare our method with simple pretraining with RGB/Flow modalities. Table~\ref{table:domain_gen} shows the performance of simple RGB/Flow pretraining and our feature-supervised method on NTU-120\_minus\_60 and  Kinetics-400  source datasets. It is evident from the table that for both datasets our method outperforms simple pretraining with RGB and Flow modalities by a considerable margin especially when using limited amounts of labeled examples during fine-tuning.
 At the same time, we observe that relying on NTU-120\_minus\_60 as source dataset performs better than the Kinetics-400, which is expected because it is more similar in domain to our target dataset, see also Fig.~\ref{fig:datasets}.
Hence, it provides features which are easier to match during transfer. Our approach is agnostic to the source dataset for  knowledge transfer, but the more similar the domain between source and target dataset, the better the action classification results.

%
%
%
%
%
%

\begin{table}[] 
 \centering
   \resizebox{1.0\columnwidth}{!}{
 \begin{tabular}{lcccc}
 \toprule
  & \multicolumn{3}{c}{\textbf{Target-Modality: Depth}}\\ \cmidrule(lr){2-4}
  \textbf{Method \& Source Domain} &  20 per-class &50 per-class & 100 per-class \\
  \midrule
From-scratch&  	{11.00}\scriptsize{$\pm$2.0} & {33.51}\scriptsize{$\pm$0.4} &{54.10}\scriptsize{$\pm$0.5} \\ 

Flow-pretraining (Kinetics) &  	{23.68}\scriptsize{$\pm$2.0} & {41.95}\scriptsize{$\pm$1.0}  & {54.45}\scriptsize{$\pm$0.5}\\ 

RGB-pretraining (Kinetics) &  	{24.84}\scriptsize{$\pm$2.0} & {42.96}\scriptsize{$\pm$1.0} & {55.05}\scriptsize{$\pm$0.5}\\

Flow-pretraining (NTU) &  {24.34}\scriptsize{$\pm$1.0} & {53.72}\scriptsize{$\pm$0.8}  & {64.88}\scriptsize{$\pm$0.4}\\ 
RGB-pretraining (NTU) &  	{41.55}\scriptsize{$\pm$1.0} & {57.20}\scriptsize{$\pm$0.8} & {66.41}\scriptsize{$\pm$0.4}\\ 

\midrule

  RGB-feature-supervised (Kinetics)  &  	{33.31}\scriptsize{$\pm$1.0} & {47.38}\scriptsize{$\pm$0.5} &{55.10}\scriptsize{$\pm$0.5} \\ 
 
  Flow-feature-supervised (Kinetics)  &  	{52.17}\scriptsize{$\pm$1.0} & {59.51}\scriptsize{$\pm$0.6} &{64.14}\scriptsize{$\pm$0.5} \\ 
 
 RGB-feature-supervised (NTU)&  	{63.05}\scriptsize{$\pm$1.0} & {66.71}\scriptsize{$\pm$0.6} &{68.64}\scriptsize{$\pm$0.3} \\ 
  
   Flow-feature-supervised (NTU) &   {68.43}\scriptsize{$\pm$0.2} & {71.53}\scriptsize{$\pm$0.1} &{73.43}\scriptsize{$\pm$0.3} \\ 
 \bottomrule
 \end{tabular}}
 \caption{\textbf{\textit{Domain Generalization.}}  Performance of the depth student under varying source domains. From-scratch indicates training directly on the target modality without any pretraining. Pretraining indicates pretraining on RGB/Flow modality of the source dataset and then directly fine-tuned with labeled depth maps from the target dataset. Feature-supervised indicates the pretraining via our cross-modal transfer with the RGB/Flow trained on a source dataset as the teacher network, followed by fine-tuning with the labeled depth maps from the target dataset. 
 All models are evaluated on the cross-view validation set of NTU RGB+D. Our method outperforms both training from scratch as well as simple pretraining. Also, the teacher network from a more similar domain provides better transfer features.
} \label{table:domain_gen}
\end{table} 

\subsection{Modality Generalization} 
We now change the student modality to 3D-skeleton data to assess to what extent our method generalizes over modalities. As discussed in section~\ref{subsec:vtov},  the video-to-video level strategy is needed for  knowledge transfer to 3D-skeleton data.  Table~\ref{table:modailty_gen} shows that our feature-supervised knowledge transfer improves the performance for the 3D-skeleton modality as well, especially for limited amounts of labeled data. 
Again we observe the optical-flow from a similar domain (NTU-120\_minus\_60) acts as the best source modality for transfer. We also observe that the performance increase is not as good as the transfer to depth maps (compare with Table~\ref{table:domain_gen}). This is because completely dissimilar modalities (image-based RGB and optical-flow \textit{vs.} joint-based 3D-Skeleton poses) and different network architectures (a 3D-CNN teacher and a graph-CNN student) are involved in this transfer. This makes it harder to match the features as compared to depth maps (image-based and 3D-CNN student). In summary, our method also generalizes to a much more difficult target modality such as 3D-skeleton data.

\begin{table}[] 
 \centering
  \resizebox{0.99\columnwidth}{!}{
 \begin{tabular}{lcccc}
 \toprule
  & \multicolumn{3}{c}{\textbf{Target-Modality: 3D-skeleton}}\\ \cmidrule(lr){2-4}
  \textbf{Source-Modality \& Domain} &  20 per-class &50 per-class & 100 per-class \\
  \midrule
From-scratch&  	{33.00}\scriptsize{$\pm$3.0} & {50.11}\scriptsize{$\pm$2.0} &{67.50}\scriptsize{$\pm$1.5} \\ 
Kinetics-RGB &  	{52.15}\scriptsize{$\pm$1.0} & {65.86}\scriptsize{$\pm$1.0} &{74.98}\scriptsize{$\pm$0.4} \\ 
Kinetics-Flow &  	{52.39}\scriptsize{$\pm$2.0} & {66.11}\scriptsize{$\pm$0.5} &{75.46}\scriptsize{$\pm$0.2} \\ 
NTU\_120\_minus\_60-RGB&  	{57.53}\scriptsize{$\pm$1.5} & {70.30}\scriptsize{$\pm$0.5} &{77.95}\scriptsize{$\pm$0.5} \\ 
NTU\_120\_minus\_60-Flow&   {58.57}\scriptsize{$\pm$1.5} & {71.11}\scriptsize{$\pm$0.6} &{78.59}\scriptsize{$\pm$0.4} \\ 
 \bottomrule
 \end{tabular}}
 \caption{\textbf{\textit{Modality Generalization.}} Performance of the 3D-skeleton student as the target modality. From-scratch indicates training the 3D-skeleton network directly on the target modality. The video-to-video strategy is used for the transfer. All models are evaluated the on cross-view validation set of NTU RGB+D. Our method generalises to a difficult 3D-skeleton target modality.
} \label{table:modailty_gen}
\end{table}

\subsection{Task Generalization} 
In this experiment we evaluate our method for the task of temporal action detection where the goal is to predict the start and end locations of multiple activities in a long untrimmed depth video, along with the action classes. The knowledge transfer step remains the same as before, however, during fine-tuning the student network is now optimized for the task of action detection using (a limited amount of) labeled examples. As before, the unlabeled modality pairs from the training set of NTU RGB+D are used for the feature-supervised knowledge transfer. For fine-tuning we now sample from the labeled training set of PKU-MMD.
 
\subsubsection{Action Detection Student Network} We rely on the R-C3D network \cite{rc3d} for this task with depth maps as the input. The architecture contains a backbone network which is connected to a region proposal network and a classification network. The student network is first trained for the knowledge transfer separately as before, and, then acts as the backbone network in the R-C3D framework during fine-tuning. We follow the same training procedure for the knowledge transfer as described in section~\ref{subsec:implementation}.  For fine-tuning, the whole R-C3D framework is trained with all the hyperparameters from \cite{rc3d}. We train the network for a total of 8 epochs with a batch size of 4. The learning rate is initialized to 0.0001 and decreased to 0.00001 for the last 2 epochs. For inference and evaluation we follow the setting suggested in \cite{rc3d}.
 
\subsubsection{Action Detection Results}  Table~\ref{table:detection}  shows  the mean average precision (mAP at an IoU threshold of 0.5) on PKU-MMD for varying amounts of labeled training data (a quarter, half and all available training examples).  We observe for all three splits there is a considerable gap in performance as compared to training from scratch as well as with simple RGB/Flow based pretraining, especially for the smallest split making cross-modal feature-supervised transfer beneficial when only limited amounts of start, end and class labels for the target modality are available.  Thus, we can also use our pre-training strategy to reduce the number of labeled examples without any drastic drop in performance. 
Finally, we again observe the optical-flow  proves to be the best modality for transfer.
 
%

\begin{table}[t!] 
 \centering
   \resizebox{0.99\columnwidth}{!}{
 \begin{tabular}{lcccc}
 \toprule
  & \multicolumn{3}{c}{\textbf{Target-Modality: Depth}}\\ \cmidrule(lr){2-4}
  \textbf{ Method } &  $1/4$ train-set &$1/2$ train-set &  entire train-set\\
  \midrule
 From-Scratch  &  52.85 & 66.45 & 73.39  \\
 RGB-pretraining   &  70.57 & 79.61 & 81.67  \\
  Flow-pretraining  &  73.68 & 81.21 & 81.72  \\
 \midrule
 RGB-feature-supervised & 78.89 & 82.73 & 85.95 \\
 Flow-feature-supervised & 79.87 & 84.85 & 86.81 \\
 %
 \bottomrule
 \end{tabular}}
 \caption{\textbf{\textit{Task Generalization.}}
 Temporal action detection results (mAP@IoU=0.5) on PKU-MMD from depth maps using an R-C3D network. From-scratch indicates training the action detection network directly on the target modality without any pretraining. Pretraining indicates the action detection backbone is pretrained with RGB/Flow modality on NTU-120\_minus\_60 dataset and then directly finetuned with the labeled untrimmed depth maps. Feature-supervised  indicates the backbone network is pretrained via our cross-modal transfer with RGB/Flow model trained on NTU-120\_minus\_60 as the teacher network, followed by fine-tuning with the labeled untrimmed depth maps.
 We use a quarter (235), half (470) or the entire (942)  video train set for fine-tuning. All methods are evaluated on the validation set of the cross-subject split of PKU-MMD. Feature-supervised transfer outperforms both training from scratch and simple pretraining for the task of  temporal action detection too.
} \label{table:detection}
\end{table} 

\begin{figure}[t!]
\begin{minipage}[b]{1.0\linewidth}
  \centering
  \centerline{\includegraphics[width=8cm]{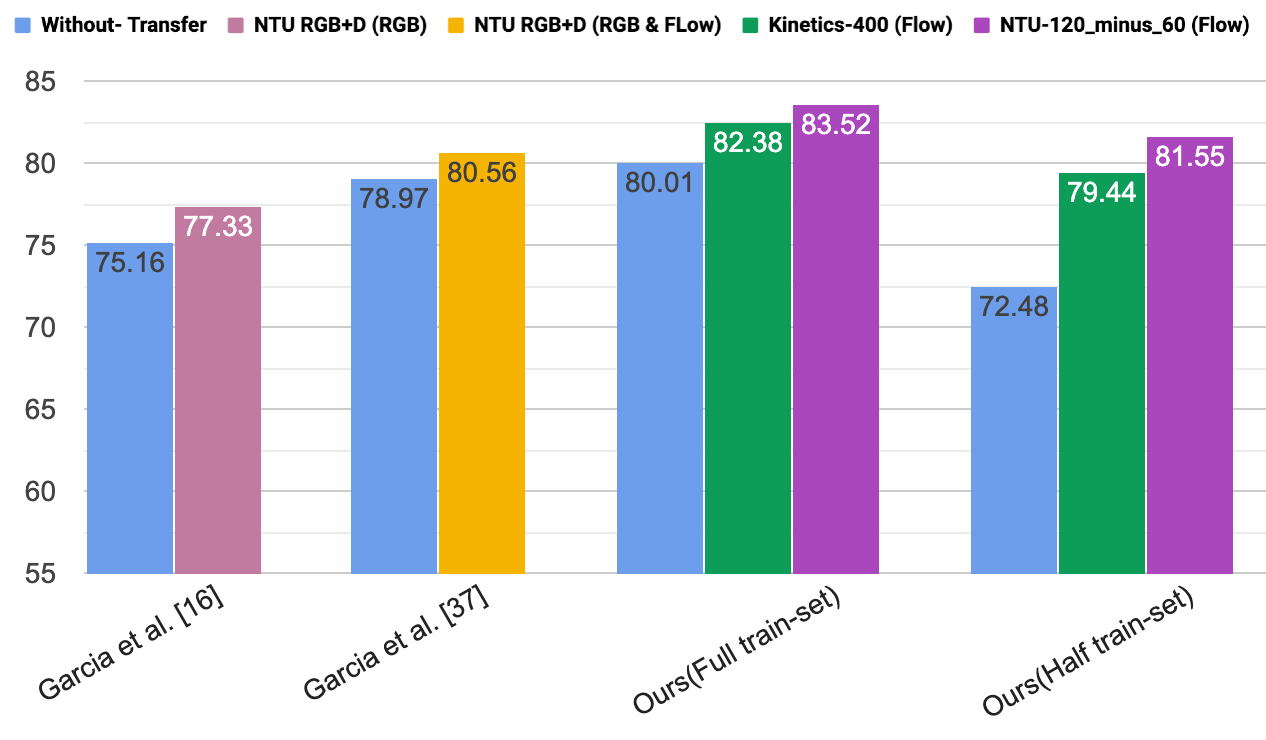}}
\end{minipage}
\caption{\textbf{\textit{Comparison with the state-of-the-art}} for depth based action recognition on NTU RGB+D for the cross-view split. Without-Transfer indicates using only depth maps from the training set.
For the rest, each column represents performance of respective method by distilling knowledge from one or more RGB action models trained on a source dataset (as shown in legend).  Each method utilizes full-training set of labeled depth maps from the training set of NTU RGB+D, except the last one that uses only half of the same train set. Our method boosts the performance for depth modality while distilling from  a different source dataset. Also, when we reduce the number of labeled examples for the depth modality by half, our method has no drastic decrease in performance.
}
\label{fig:state_of_the_art}
\end{figure}

\subsection{Comparison with the state-of-the-art} 
We now compare our method with the state of the art methods for depth-based action recognition that rely on additional modalities for transferring knowledge during training, along with using the labeled depth examples.  
Fig.~\ref{fig:state_of_the_art} shows the results for action classification from depth maps on the NTU RGB+D dataset. 
The method by Garcia \etal \cite{garci_md} uses a four-step process that relies on labeled RGB-depth pairs from NTU RGB+D  to distill knowledge from the RGB stream to the depth stream. They achieve  a 2\% improvement over their baseline (\ie, training a depth stream without any knowledge transfer). Similarly, another method by Garcia \etal \cite{garcia2019dmcl} relies on labeled RGB-optical-flow-depth triplets  from NTU RGB+D to train three networks together, such that, the RGB and the optical-flow streams are used to transfer knowledge to the depth stream during training. They achieve a boost of 1.5\%  over their baseline.  Both methods use the full training set of labeled depth maps and other source modalities (also with labels) from NTU RGB+D  to achieve this performance.

Our method relies on unlabeled RGB-depth or optical-flow-depth pairs from NTU RGB+D and pre-trained action models from other RGB datasets (like Kinetics-400 and NTU-120\_minus\_60)  to distill knowledge to a depth stream. Followed by fine-tuning with labeled depth examples from the NTU RGB+D training set.
For the full train set, we achieve a boost of around 2\% and 3\% over our baseline by transferring from the optical-flow teachers trained on Kinetics-400 and NTU-120\_minus\_60, respectively. We further show by reducing the number of labeled depth examples from NTU RGB+D for fine-tuning by half, the performance drop is small. Thus, in this way our method can act as pre-training source for reducing the number of labeled examples for non-RGB modalities such as depth maps.
In other words, our method is particularly useful for the scenario where labeled examples for the target dataset are  scarce and available in the target action modality only.

\section{Conclusion}
In this work, we presented a method to train action models for a non-RGB target modality, such as depth maps or 3D-skeletons, by extracting knowledge from a large-scale action labeled RGB dataset. Unlabeled pairs of the RGB and target modality are leveraged for cross-modal knowledge transfer by feature-supervision. 
Our extensive evaluation showed that we can use pre-trained RGB action models (particularly optical-flow from a more similar domain) to  transfer knowledge to  recognize and detect new actions from  other action modalities like depth maps and 3D-skeleton sequences.  In conclusion, we showed how large RGB action datasets can be used as valuable pre-training source for other non-RGB action datasets with limited labeled examples.

\section{ACKNOWLEDGEMENTS}
This work is part of the research programme Perspectief EDL with project number P16-25 project 3, which is financed by the Dutch Research Council (NWO) domain Applied and Engineering/ Sciences (TTW).

\bibliographystyle{IEEEtran}
\bibliography{IEEEexample}

\begin{thebibliography}{10}
\providecommand{\url}[1]{#1}
\csname url@samestyle\endcsname
\providecommand{\newblock}{\relax}
\providecommand{\bibinfo}[2]{#2}
\providecommand{\BIBentrySTDinterwordspacing}{\spaceskip=0pt\relax}
\providecommand{\BIBentryALTinterwordstretchfactor}{4}
\providecommand{\BIBentryALTinterwordspacing}{\spaceskip=\fontdimen2\font plus
\BIBentryALTinterwordstretchfactor\fontdimen3\font minus
  \fontdimen4\font\relax}
\providecommand{\BIBforeignlanguage}[2]{{%
\expandafter\ifx\csname l@#1\endcsname\relax
\typeout{** WARNING: IEEEtran.bst: No hyphenation pattern has been}%
\typeout{** loaded for the language `#1'. Using the pattern for}%
\typeout{** the default language instead.}%
\else
\language=\csname l@#1\endcsname
\fi
#2}}
\providecommand{\BIBdecl}{\relax}
\BIBdecl

\bibitem{slow_fast}
C.~Feichtenhofer, H.~Fan, J.~Malik, and K.~He, ``Slowfast networks for video
  recognition,'' in \emph{ICCV}, 2019.

\bibitem{kinetics}
J.~{Carreira} and A.~{Zisserman}, ``Quo vadis, action recognition? a new model
  and the kinetics dataset,'' in \emph{CVPR}, 2017.

\bibitem{hara3dcnns}
K.~Hara, H.~Kataoka, and Y.~Satoh, ``Can spatiotemporal 3d cnns retrace the
  history of 2d cnns and imagenet?'' in \emph{CVPR}, 2018.

\bibitem{Depthmotionmap}
X.~Yang, C.~Zhang, and Y.~Tian, ``Recognizing actions using depth motion
  maps-based histograms of oriented gradients,'' in \emph{ACM Multimedia},
  2012.

\bibitem{3ddepth}
Z.~Liu, C.~Zhang, and Y.~Tian, ``3d-based deep convolutional neural network for
  action recognition with depth sequences,'' in \emph{Image Vision Comput.},
  vol.~55, no.~P2, Nov. 2016, pp. 93--100.

\bibitem{DBLP:deep_depth}
P.~Wang, W.~Li, Z.~Gao, J.~Zhang, C.~Tang, and P.~Ogunbona, ``Deep
  convolutional neural networks for action recognition using depth map
  sequences,'' \emph{CoRR}, vol. abs/1501.04686, 2015.

\bibitem{2sagcn2019cvpr}
L.~Shi, Y.~Zhang, J.~Cheng, and H.~Lu, ``Two-stream adaptive graph
  convolutional networks for skeleton-based action recognition,'' in
  \emph{CVPR}, 2019.

\bibitem{Shi_2019_CVPR}
S.~Lei, Z.~Yifan, C.~Jian, and L.~Hanqing, ``Skeleton-based action recognition
  with directed graph neural networks,'' in \emph{CVPR}, 2019.

\bibitem{Wu_2019_ICCV}
C.~Wu, X.-J. Wu, and J.~Kittler, ``Spatial residual layer and dense connection
  block enhanced spatial temporal graph convolutional network for
  skeleton-based action recognition,'' in \emph{ICCV Workshops}, 2019.

\bibitem{Sports1M}
A.~Karpathy, G.~Toderici, S.~Shetty, T.~Leung, R.~Sukthankar, and L.~Fei-Fei,
  ``Large-scale video classification with convolutional neural networks,'' in
  \emph{CVPR}, 2014.

\bibitem{caba2015activitynet}
B.~G. Fabian Caba~Heilbron, Victor~Escorcia and J.~C. Niebles, ``Activitynet: A
  large-scale video benchmark for human activity understanding,'' in
  \emph{CVPR}, 2015.

\bibitem{ucf-101}
K.~Soomro, A.~R. Zamir, and M.~Shah, ``Ucf101: A dataset of 101 human action
  classes from videos in the wild,'' in \emph{CRCV-TR-12-01}, 2012.

\bibitem{hmdb}
H.~Kuhne, H.~Jhuang, E.~Garrote, T.~Poggio, and T.~Serre, ``Hmdb: A large video
  database for human motion recognition,'' in \emph{ICCV}, 2011.

\bibitem{THUMOS14}
H.~Idrees, A.~R. Zamir, Y.-G. Jiang, A.~Gorban, I.~Laptev, R.~Sukthankar, and
  M.~Shah, ``The {THUMOS} challenge on action recognition for videos in the
  wild,'' \emph{CVIU}, 2017.

\bibitem{kd}
G.~Hinton, O.~Vinyals, and J.~Dean, ``Distilling the knowledge in a neural
  network,'' in \emph{NeurIPS Workshop}, 2015.

\bibitem{garci_md}
N.~C. Garcia, P.~Morerio, and V.~Murino, ``Modality distillation with multiple
  stream networks for action recognition,'' in \emph{ECCV}, 2018.

\bibitem{garcia2018admd}
N.~Garcia, P.~Morerio, and V.~Murino, ``Learning with privileged information
  via adversarial discriminative modality distillation,'' \emph{in PAMI}, 2019.

\bibitem{crasto2019mars}
N.~Crasto, P.~Weinzaepfel, K.~Alahari, and C.~Schmid, ``{MARS: Motion-Augmented
  RGB Stream for Action Recognition},'' in \emph{CVPR}, 2019.

\bibitem{fmthoker_icip}
F.~M. Thoker and J.~Gall, ``Cross-modal knowledge distillation for action
  recognition,'' in \emph{ICIP}, 2019.

\bibitem{c3d}
D.~Tran, L.~Bourdev, R.~Fergus, L.~Torresani, and M.~Paluri, ``Learning
  spatiotemporal features with 3d convolutional networks,'' in \emph{ICCV},
  2015.

\bibitem{r2d+1}
D.~{Tran}, H.~{Wang}, L.~{Torresani}, J.~{Ray}, Y.~{LeCun}, and M.~{Paluri},
  ``A closer look at spatiotemporal convolutions for action recognition,'' in
  \emph{CVPR}, 2018.

\bibitem{TSN2016ECCV}
L.~Wang, Y.~Xiong, Z.~Wang, Y.~Qiao, D.~Lin, X.~Tang, and L.~{Val Gool},
  ``Temporal segment networks: Towards good practices for deep action
  recognition,'' in \emph{ECCV}, 2016.

\bibitem{SimonyanZ14}
K.~Simonyan and A.~Zisserman, ``Two-stream convolutional networks for action
  recognition in videos,'' in \emph{NeurIPS}, 2014.

\bibitem{feichtenhofer2016convolutional}
C.~Feichtenhofer, A.~Pinz, and A.~Zisserman, ``Convolutional two-stream network
  fusion for video action recognition,'' in \emph{CVPR}, 2016.

\bibitem{zhao2019twoinone}
J.~Zhao and C.~G.~M. Snoek, ``Dance with flow: Two-in-one stream action
  detection,'' in \emph{CVPR}, 2019.

\bibitem{stop}
A.~W. Vieira, E.~R. Nascimento, G.~L. Oliveira, Z.~Liu, and M.~F.~M. Campos,
  ``Stop: Space-time occupancy patterns for 3d action recognition from depth
  map sequences,'' in \emph{Pattern Recognition, Image Analysis, Computer
  Vision, and Applications}.\hskip 1em plus 0.5em minus 0.4em\relax Berlin,
  Heidelberg: Springer Berlin Heidelberg, 2012, pp. 252--259.

\bibitem{stgcn2018aaai}
S.~Yan, Y.~Xiong, and D.~Lin, ``Spatial temporal graph convolutional networks
  for skeleton-based action recognition,'' in \emph{AAAI}, 2018.

\bibitem{li2019rf}
T.~Li, L.~Fan, M.~Zhao, Y.~Liu, and D.~Katabi, ``Making the invisible visible:
  Action recognition through walls and occlusions,'' in \emph{ICCV}, 2019.

\bibitem{emotion}
S.~Albanie, A.~Nagrani, A.~Vedaldi, and A.~Zisserman, ``Emotion recognition in
  speech using cross-modal transfer in the wild,'' in \emph{ACM Multimedia},
  2018.

\bibitem{wall_pose}
M.~Zhao, T.~Li, M.~Abu~Alsheikh, Y.~Tian, H.~Zhao, A.~Torralba, and D.~Katabi,
  ``Through-wall human pose estimation using radio signals,'' in \emph{CVPR},
  2018.

\bibitem{hoffman2016learning}
J.~Hoffman, S.~Gupta, and T.~Darrell, ``Learning with side information through
  modality hallucination,'' in \emph{CVPR}, 2016.

\bibitem{gupta2015cross}
S.~Gupta, J.~Hoffman, and J.~Malik, ``Cross modal distillation for supervision
  transfer,'' in \emph{{CVPR}}, 2016.

\bibitem{object_detect}
G.~Chen, W.~Choi, X.~Yu, T.~Han, and M.~Chandraker, ``Learning efficient object
  detection models with knowledge distillation,'' in \emph{NIPS}, 2017.

\bibitem{video_caption_1}
B.~Zhang, H.~Hu, and F.~Sha, ``Cross-modal and hierarchical modeling of video
  and text,'' in \emph{ECCV}, 2018.

\bibitem{video_caption_2}
X.~Wang, Y.-F. Wang, and W.~Y. Wang, ``Watch, listen, and describe: Globally
  and locally aligned cross-modal attentions for video captioning,'' in
  \emph{NAACL}, 2018.

\bibitem{d3d}
J.~Stroud, D.~Ross, C.~Sun, J.~Deng, and R.~Sukthankar, ``D3d: Distilled 3d
  networks for video action recognition,'' in \emph{WACV}, 2020.

\bibitem{garcia2019dmcl}
N.~C. Garcia, S.~A. Bargal, V.~Ablavsky, P.~Morerio, V.~Murino, and
  S.~Sclaroff, ``Dmcl: Distillation multiple choice learning for multimodal
  action recognition,'' \emph{arXiv preprint arXiv:1912.10982}, 2019.

\bibitem{sayed:GCPR:2018}
N.~Sayed, B.~Brattoli, and B.~Ommer, ``Cross and learn: Cross-modal
  self-supervision,'' in \emph{GCPR}, 2018.

\bibitem{Shahroudy_2016_NTURGBD}
A.~Shahroudy, J.~Liu, T.-T. Ng, and G.~Wang, ``Ntu rgb+d: A large scale dataset
  for 3d human activity analysis,'' in \emph{CVPR}, 2016.

\bibitem{liu2017pku}
L.~Chunhui, H.~Yueyu, L.~Yanghao, S.~Sijie, and L.~Jiaying, ``Pku-mmd: A large
  scale benchmark for continuous multi-modal human action understanding,''
  \emph{ACM Multimedia workshop}, 2017.

\bibitem{Liu_2019_NTURGBD120}
J.~Liu, A.~Shahroudy, M.~Perez, G.~Wang, L.-Y. Duan, and A.~C. Kot, ``Ntu rgb+d
  120: A large-scale benchmark for 3d human activity understanding,'' \emph{in
  PAMI}, 2019.

\bibitem{rc3d}
H.~Xu, A.~Das, and K.~Saenko, ``R-c3d: region convolutional 3d network for
  temporal activity detection,'' in \emph{ICCV}, 2017.

\end{thebibliography}
\end{document}